\documentclass[letterpaper, 10 pt, conference]{ieeeconf}

\IEEEoverridecommandlockouts

\usepackage{graphicx}
\usepackage{tabularx}

\usepackage{cite}
\usepackage{amsmath,amssymb,amsfonts}
\usepackage{algorithmic}
\usepackage{textcomp}
\usepackage{pifont} %

\newcommand{\cmark}{\textcolor{green!60!black}{\ding{51}}} %
\newcommand{\xmark}{\textcolor{red}{\ding{55}}}         %

\usepackage{balance}
\usepackage{comment}
\usepackage[ruled,vlined]{algorithm2e}
\usepackage{caption}

\usepackage{multirow}
\usepackage{array}
\usepackage{subcaption}
\usepackage{arydshln}
\usepackage{tensor}
\newcommand{\eg}{\emph{e.g.},}
\newcommand{\ie}{\emph{i.e.},}

\usepackage{hyperref}
\usepackage{float}
\usepackage{rotating}
\usepackage{nicematrix}  %
\usepackage{mathtools}
\usepackage{amsmath}
\usepackage{pifont}
\usepackage{cuted}
\usepackage{dblfloatfix}
\usepackage{gensymb}
\usepackage[ruled,vlined]{algorithm2e}
\usepackage{titlesec}
\usepackage{amssymb}
\usepackage{graphicx}
\let\labelindent\relax
\usepackage{enumitem}
\titlespacing*{\section}{0pt}{.5em}{0pt}
\usepackage{cuted}
\definecolor{LightCyan}{rgb}{0.88,1,0.88}
\definecolor{linear_color}{RGB}{220,223,240}
\definecolor{gray_bbox_color}{RGB}{243,243,244}

\definecolor{rebuttal}{rgb}{0,0,1}

\def\eqref#1{Eq.~(\ref{#1})}

\def\dsname{\textit{WildCross}}

\definecolor{testteal}{RGB}{204, 236, 239}

\newcommand{\coolname}{WildCross}
\titlespacing*{\subsection}{0pt}{0.5ex}{0.5ex}

\begin{document}

\bstctlcite{IEEEexample:BSTcontrol}

\title{\LARGE \bf \coolname: A Cross-Modal Large Scale Benchmark for Place Recognition and Metric Depth Estimation in Natural Environments \vspace{-3mm}
}
\author{Joshua Knights$^{1,2}$, Joseph Reid$^{1}$, Kaushik Roy$^{1}$, David Hall$^{1}$, Mark Cox$^{1}$, Peyman Moghadam$^{1,2}$
\thanks{$^{1}$ CSIRO Robotics, Data61, CSIRO, Australia. E-mail: {\tt\footnotesize \emph{firstname.lastname}@csiro.au}}
\thanks{$^{2}$ Queensland University of Technology (QUT), Australia.}
}

\maketitle
\begin{strip}
\centering
\vspace{-2cm}
    \captionsetup{type=figure}
    \includegraphics[width=\textwidth]{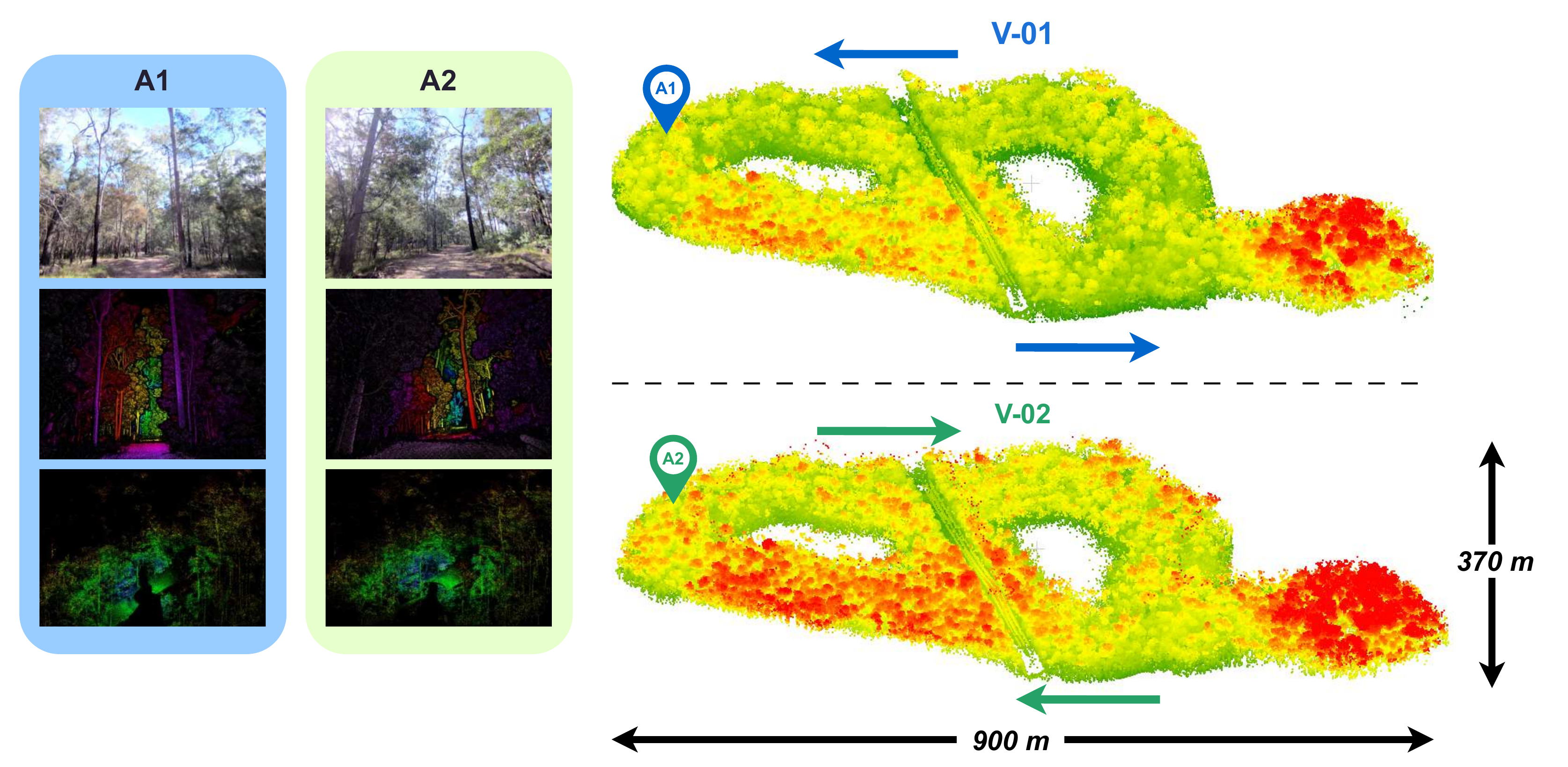}
    \captionof{figure}{The global maps of two sequences from \dsname{}. The left panels show the RGB images (top), annotated depth images (middle), and lidar submaps (bottom) at locations A1 and A2. These correspond to revisits of the same location from opposite directions across different sessions. \dsname{} presents a challenging new benchmark for cross-modal place recognition and metric depth estimation, with eight traversals covering diverse viewpoints in two large-scale forests. }%
    \label{fig:hero}
    \vspace{-2mm}
\end{strip}%

\everypar{\looseness=-1}

\begin{abstract}
Recent years have seen a significant increase in demand for robotic solutions in unstructured natural environments, alongside growing interest in bridging 2D and 3D scene understanding.  However, existing robotics datasets are predominantly captured in structured urban environments, making them inadequate for addressing the challenges posed by complex, unstructured natural settings.  To address this gap, we propose \dsname{}, a cross-modal benchmark for place recognition and metric depth estimation in large-scale natural environments.  \dsname{} comprises over 476K sequential RGB frames with semi-dense depth and surface normal annotations, each aligned with accurate 6DoF poses and synchronized dense lidar submaps. We conduct comprehensive experiments on visual, lidar, and cross-modal place recognition, as well as metric depth estimation, demonstrating the value of \dsname{} as a challenging benchmark for multi-modal robotic perception tasks.  We provide access to the code repository and dataset at \href{https://csiro-robotics.github.io/WildCross}{https://csiro-robotics.github.io/WildCross}.\looseness=-1
\end{abstract}

\section{Introduction}
\begin{table*}[t]
    \centering
    \begin{NiceTabular}{l cccc c ccc}
        \Block{2-1}{Name} 
        & \Block{1-4}{Supported Tasks} &&&& 
        \Block{1-4}{Diversity} &&& \\ 
        \cline{2-5}  \cline{7-9}
        & VPR & LPR & CMPR & Depth Est. & & ViewPoint & Temporal & Scene \\\hline
        Nordland\cite{sunderhauf2013we} & \cmark&\xmark&\xmark&\xmark&&$\star$&$\star\star\star$&$\star\star\star$\\
        RELLIS-3D\cite{jiang2021rellis} & \cmark&\cmark&\cmark&\xmark&&$\star$&$\star$&$\star$\\
        Wild-Places\cite{knights2022wild} & \xmark&\cmark&\xmark&\xmark&&$\star\star\star$&$\star\star\star$&$\star\star\star$\\
        Oxford Forest\cite{oh2024evaluation} &\xmark&\cmark&\xmark&\xmark&&$\star\star\star$&$\star\star$&$\star\star\star$\\
        BotanicGarden\cite{liu2024botanicgarden} & \cmark&\cmark&\cmark&\xmark&&$\star\star$&$\star$&$\star$\\
        \textbf{\dsname{} (Ours)} & \cmark&\cmark&\cmark&\cmark&&$\star\star\star$&$\star\star\star$&$\star\star\star$ \\
    \end{NiceTabular}
    \caption{Comparison of existing datasets in natural environments with support for relocalization.  Datasets are compared based on the range of supported tasks and their diversity across viewpoint, temporal, and scene variations between revisits.}
    \label{tab:nattydatasets}
    \vspace{-7mm}
\end{table*}

Autonomous robots are increasingly deployed in unstructured and natural environments for applications such as agriculture, environmental monitoring, and search and rescue~\cite{oliveira2021advancesagri,malladi2025digiforests,blei2025cloudtrack}. 
However, progress in robotic navigation and perception tasks remains heavily dependent on public datasets, given the high cost and logistical challenges of large-scale field trials. Benchmarks such as KITTI~\cite{geiger2013vision} and Oxford RobotCar~\cite{maddern20171} have been instrumental in advancing the field, but they are predominantly captured in structured urban or indoor settings~\cite{silberman2012indoor,geiger2013vision,maddern20171}. In contrast, natural environments are characterized by irregular terrain, dense vegetation, narrow trails, and complex occlusions, rendering existing datasets insufficient for evaluating robotic autonomy in environments where it is most urgently required.  Concurrently, the robotics and computer vision communities are placing increasing emphasis on bridging 2D and 3D scene understanding, exemplified by recent advances in learning-based 3D reconstruction~\cite{wang2024dust3r,leroy2024grounding,wang2025vggt} and cross-modal place recognition~\cite{cai2024voloc,zhao2023attention,shubodh2024lip,xu2024c2l}. To support these developments, datasets must provide accurate ground truth across both 2D and 3D modalities under the added complexity of natural scenes. To this end, we present \dsname{}, a large-scale multi-modal benchmark designed to advance cross-modal place recognition and metric depth estimation in natural environments.

We derive a novel benchmark dataset from Wild-Places~\cite{knights2022wild} by extending it in two key directions. 
First, we regenerate the original sequences with accurate camera poses, enabling large-scale training and evaluation using RGB video data alongside synchronized lidar submaps. Second, we develop an annotation pipeline that produces semi-dense depth images by combining accumulated point cloud maps with robust point visibility estimation. This enables, for the first time, reliable benchmarking of metric depth estimation in challenging natural environments alongside visual, lidar, and cross-modal place recognition. 
The resulting dataset comprises over 476K sequential high-resolution RGB frames with corresponding semi-dense depth and surface normal annotations, alongside accurate 6-DoF poses and extrinsics for synchronization to the lidar submaps in the Wild-Places dataset. Spanning eight traversals over 14 months and incorporating diverse viewpoints, \dsname{} establishes a challenging and versatile benchmark for visual, lidar, and cross-modal place recognition, as well as metric depth estimation in unstructured environments.
In summary, the contributions of this paper are as follows:\looseness=-1
\begin{enumerate}[leftmargin=*]
    \item We present \dsname{}, a large-scale multi-modal benchmark for natural environments.  \dsname{} contains over 476K sequential RGB frames with corresponding semi-dense depth annotations and surface normal annotations, each with accurate 6DoF ground truth poses.
    \item We demonstrate a new method for generating semi-dense depth and surface normal annotations for each image in our dataset through a combination of leveraging a dense accumulated point cloud map, accurate camera poses, and a robust point visibility estimation pipeline to ensure a high degree of accuracy in our depth annotations, providing a valuable benchmark for metric depth estimation in natural environments\looseness=-1
    \item We conduct extensive experiments on \dsname{} by evaluating several state-of-the-art methods for visual and cross-modal place recognition, as well as metric depth estimation. Our results show that leading methods struggle in these challenging natural environments, underscoring the difficulty of the \dsname{} benchmark and opening new opportunities for future research. %
\end{enumerate}

\section{Related Work}

\begin{figure*}[t]
    \centering
    \begin{subfigure}[b]{0.38\columnwidth}
        \centering
         \includegraphics[width=\textwidth]{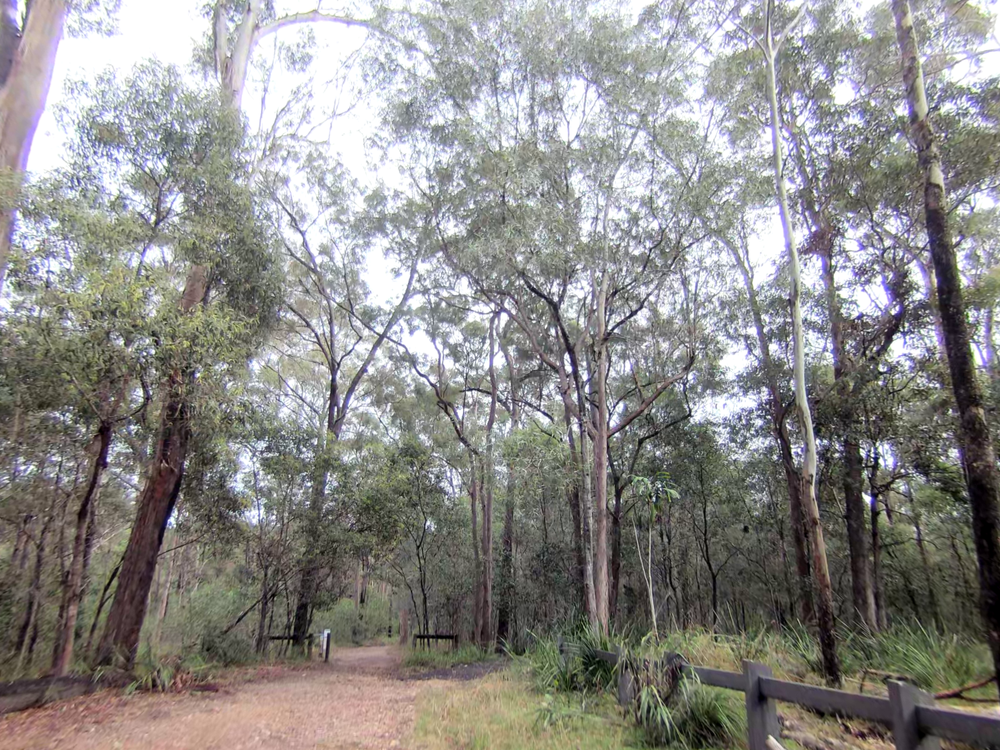}
     \end{subfigure}
    \begin{subfigure}[b]{0.38\columnwidth}
    \centering
     \includegraphics[width=\textwidth]{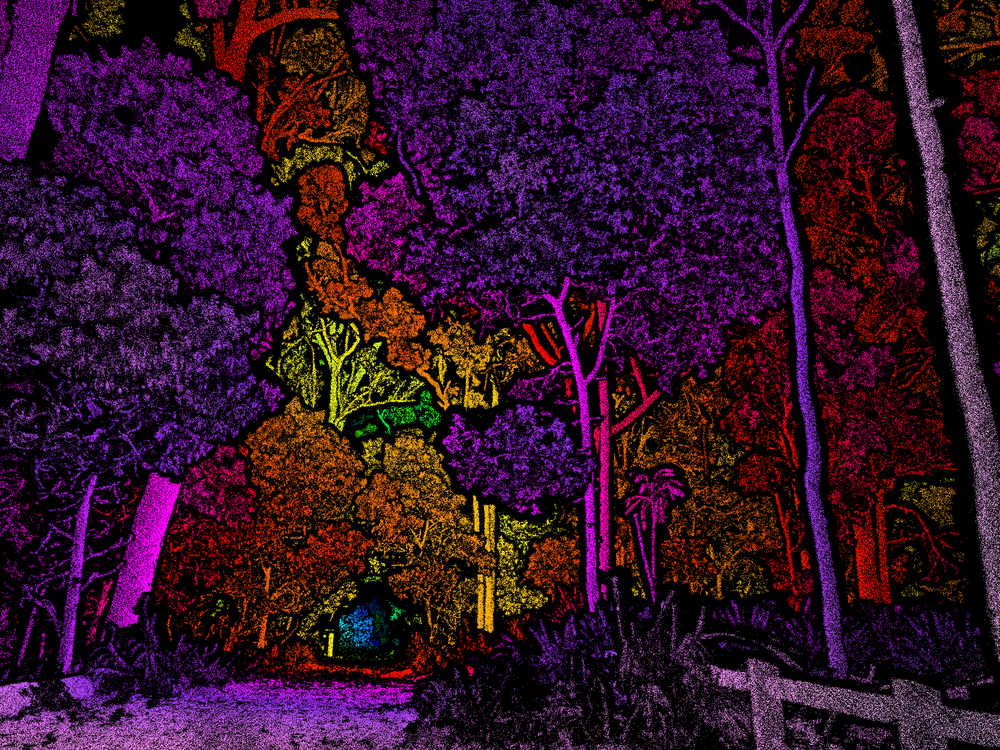}
     \end{subfigure}
    \begin{subfigure}[b]{0.38\columnwidth}
    \centering
     \includegraphics[width=\textwidth]{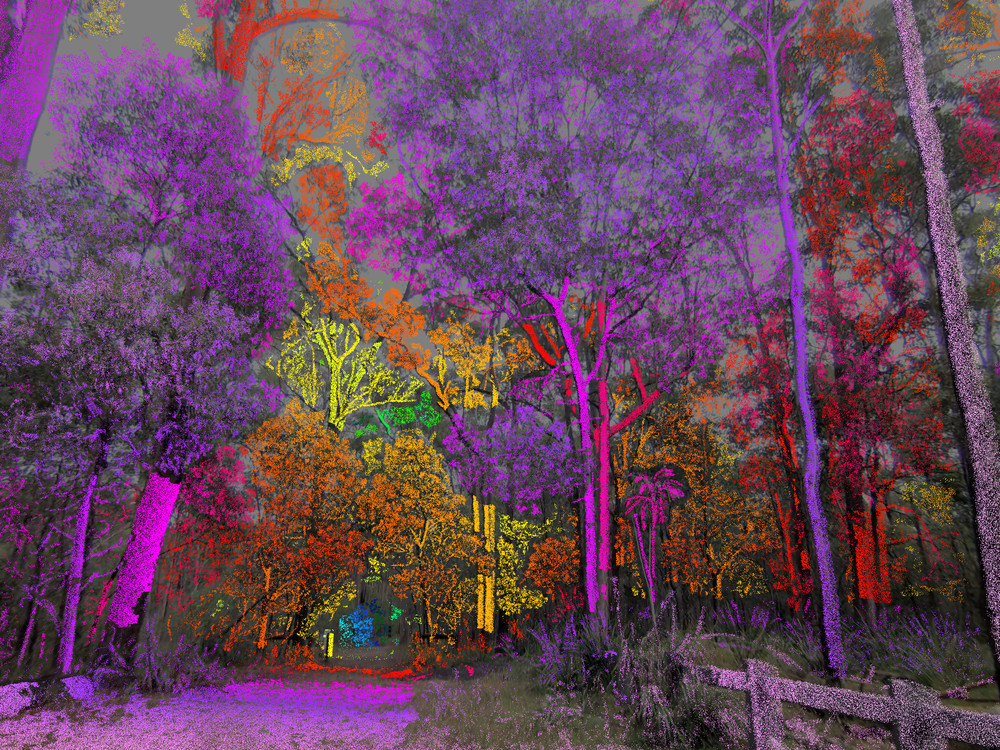}
     \end{subfigure}
    \begin{subfigure}[b]{0.38\columnwidth}
      \centering
       \includegraphics[width=\textwidth]{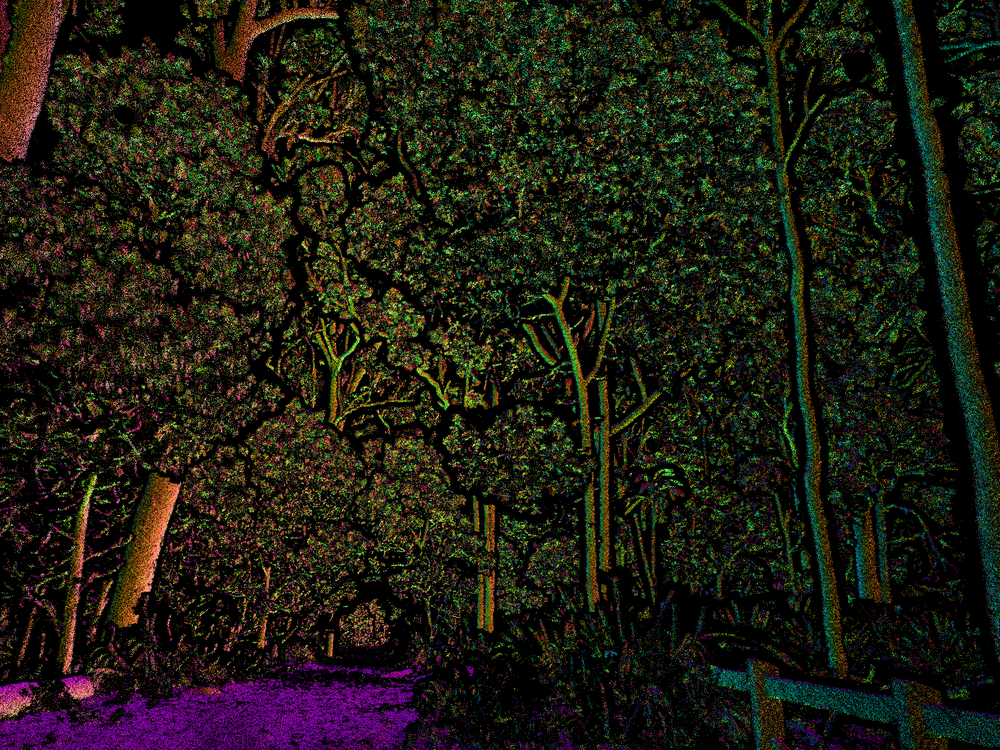}
    \end{subfigure}
    \begin{subfigure}[b]{0.38\columnwidth}
    \centering
     \includegraphics[width=\textwidth]{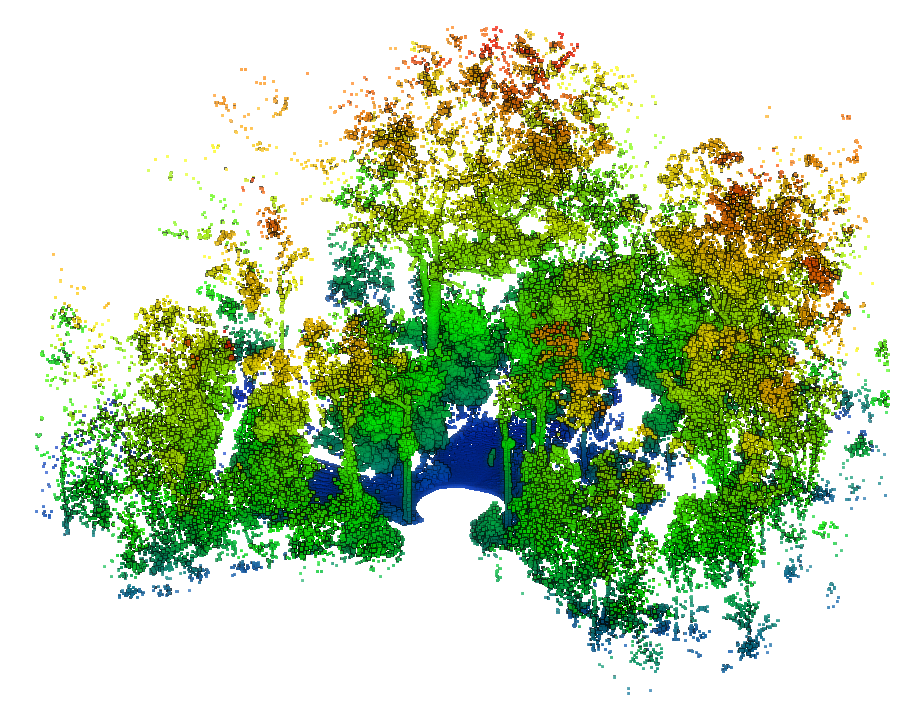}
     \end{subfigure}
     
    \vspace{1mm}
    
    \begin{subfigure}[b]{0.38\columnwidth}
        \centering
         \includegraphics[width=\textwidth]{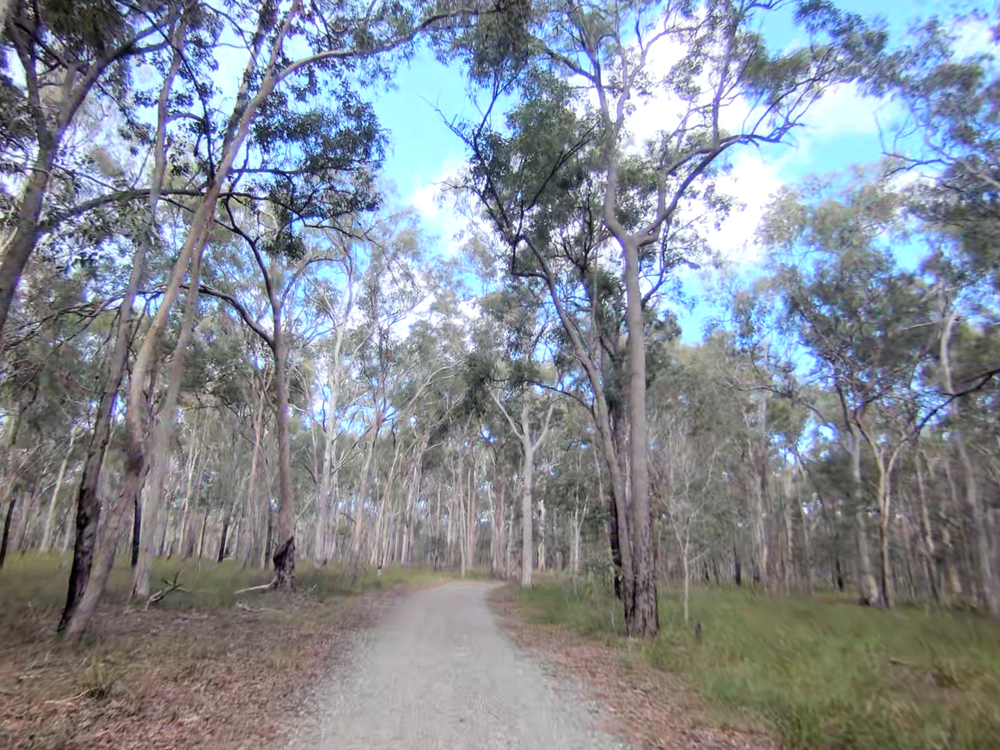}
         \caption{RGB Image}
     \end{subfigure}
    \begin{subfigure}[b]{0.38\columnwidth}
    \centering
     \includegraphics[width=\textwidth]{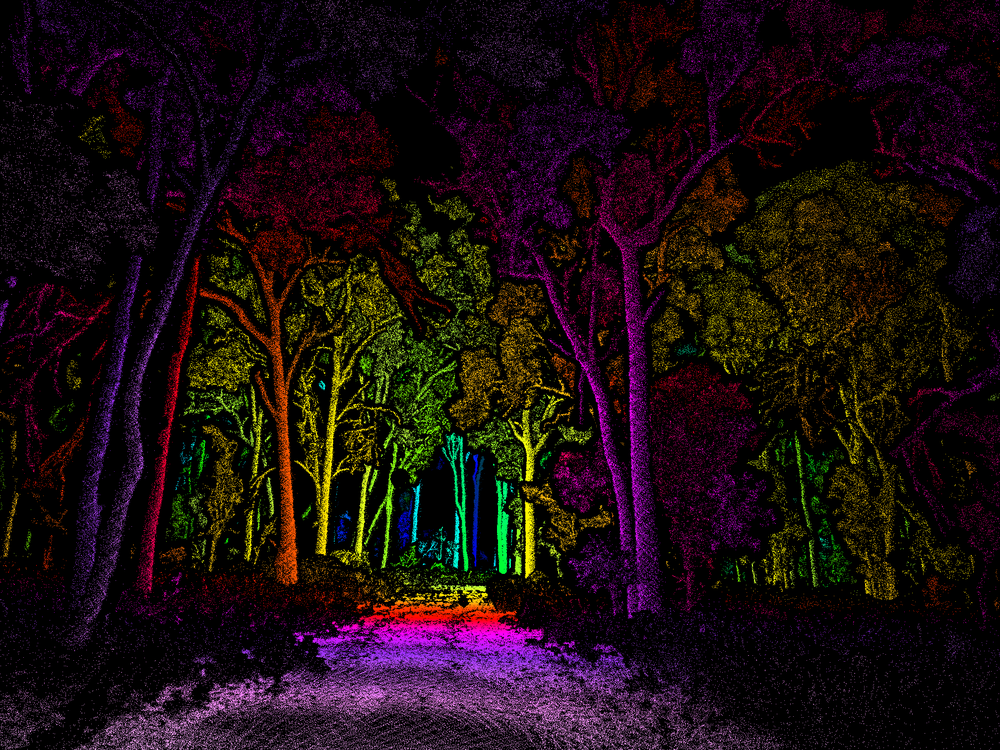}
     \caption{Depth Image}
     \end{subfigure}
    \begin{subfigure}[b]{0.38\columnwidth}
    \centering
     \includegraphics[width=\textwidth]{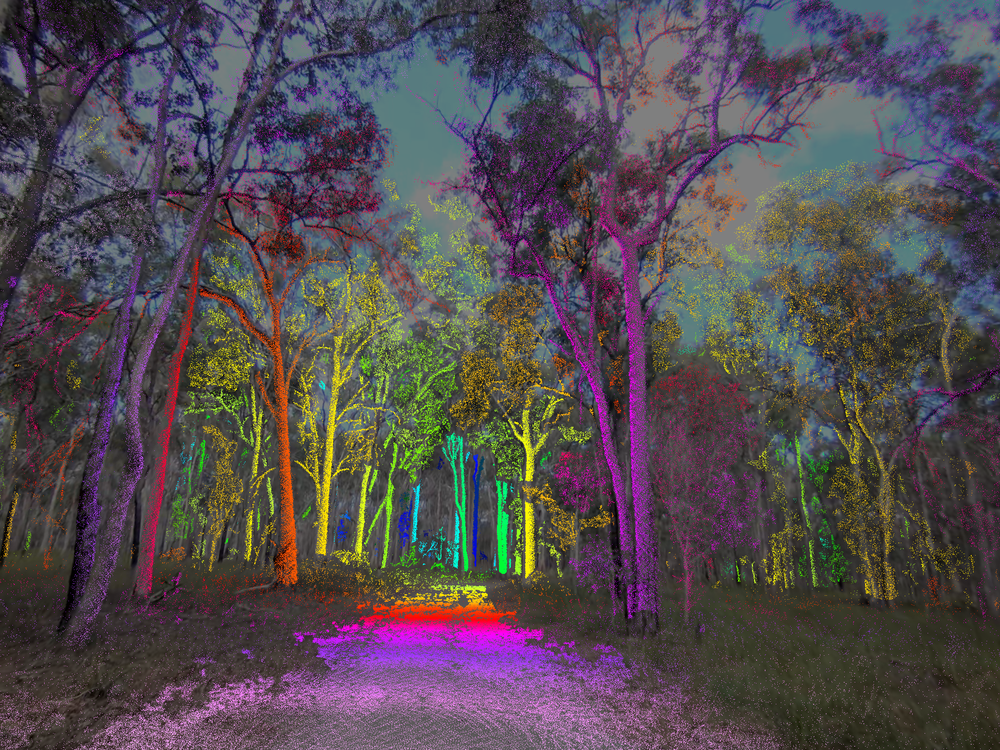}
     \caption{Depth Overlay}
     \end{subfigure}
    \begin{subfigure}[b]{0.38\columnwidth}
    \centering
     \includegraphics[width=\textwidth]{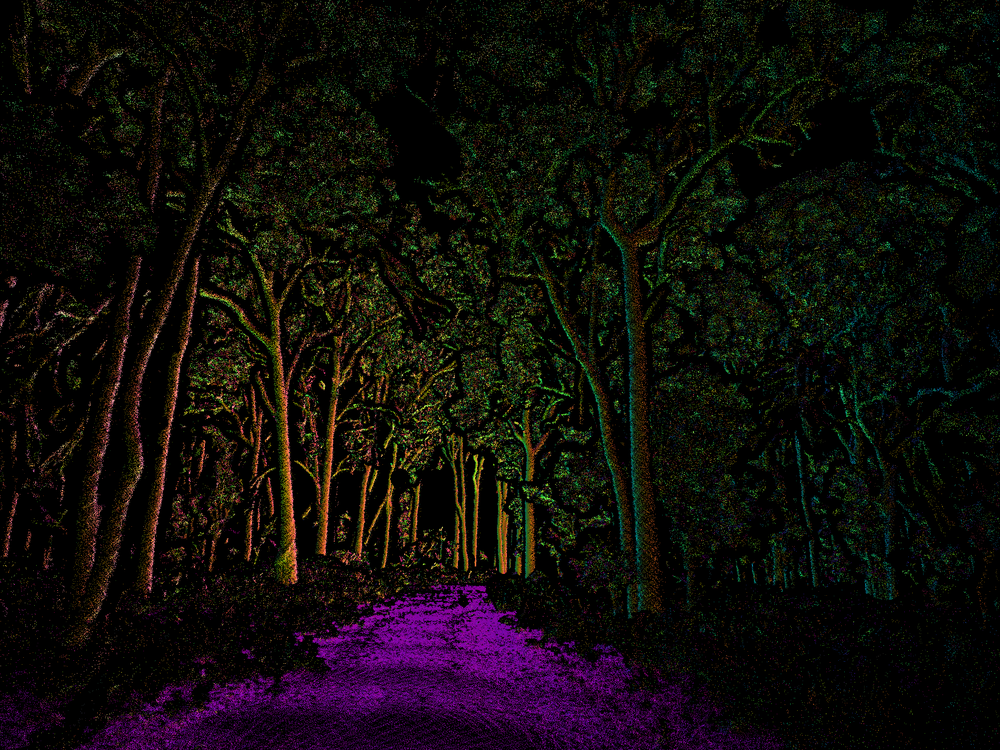}
     \caption{Surface Normal}
     \end{subfigure}
    \begin{subfigure}[b]{0.38\columnwidth}
    \centering
     \includegraphics[width=\textwidth]{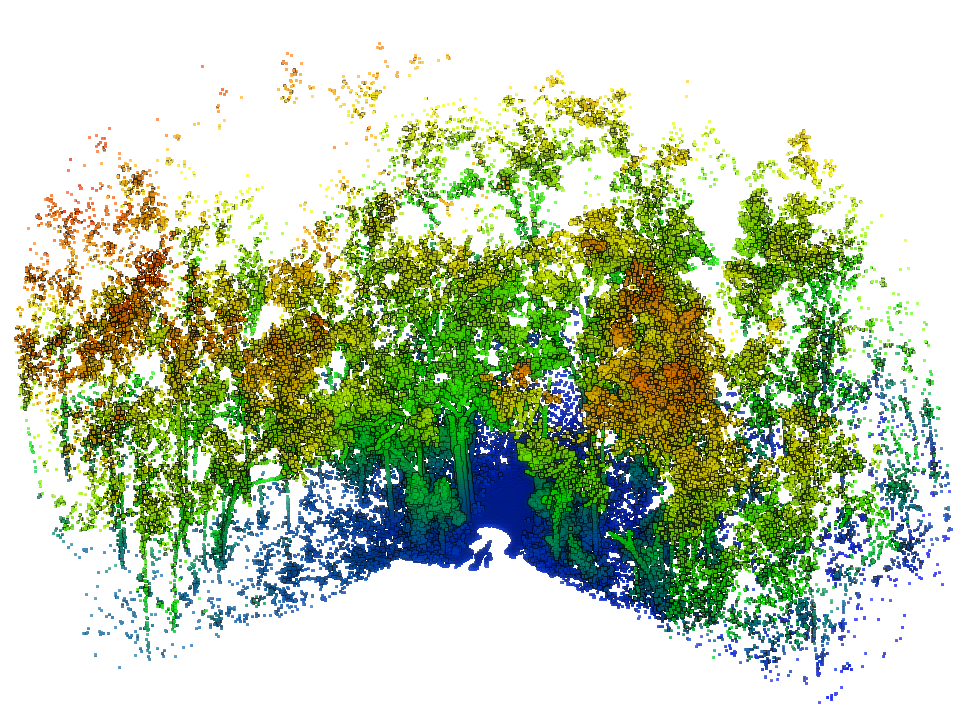}
     \caption{Lidar Submap}
     \end{subfigure}
\caption{\dsname{} overview (a) RGB Image, (b) Depth Image, (c) Depth Overlay, (d) Surface Normal, (e) Lidar Submap.}
\label{fig:dataset_contents}
\vspace{-5mm}
\end{figure*}

\subsection{Place Recognition in Natural Environments}
Place recognition is essential for the safe and reliable deployment of autonomous robots in complex environments, enabling both loop closure detection in simultaneous localization and mapping (SLAM), and re-localization in previously visited environments.
Progress in this field is largely driven by the availability of high-quality datasets for training and evaluation, particularly in VPR, where many recent state-of-the-art approaches~\cite{Izquierdo_CVPR_2024_SALAD,ali2024boq, hausler2025pair} leverage large-scale pre-training datasets such as GSV-Cities~\cite{ali2022gsv}, Google Landmarks v2~\cite{weyand2020google}, and SF-XL~\cite{Berton_CVPR_2022_CosPlace}.  
More recently, there has also been growing interest in cross-modal place recognition~ (CMPR)~\cite{cai2024voloc,zhao2023attention,shubodh2024lip,xu2024c2l, knights2025solvr}, which requires datasets with synchronized visual and lidar data for training and evaluation.\looseness=-1

Recent state-of-the-art place recognition approaches continue to demonstrate strong generalization across urban benchmarks for VPR~\cite{torii2013visual,torii201524,warburg2020mapillary}, LPR~\cite{kim2020mulran,geiger2013vision,carlevaris2016university}, and CMPR~\cite{maddern20171,geiger2013vision}. This progress has been driven in part by the growing adoption of large-scale pre-trained foundation models~\cite{Izquierdo_CVPR_2024_SALAD} and task-specific pre-training strategies~\cite{hausler2025pair}. However, even methods that benefit from such large-scale pre-training often degrade sharply under the severe domain shift between urban and natural environments. These limitations motivate the development of new large-scale datasets that capture the complexity and variability of unstructured natural environments.

Table \ref{tab:nattydatasets} provides an overview of existing datasets with support for place recognition evaluation in natural environments.  Nordland ~\cite{sunderhauf2013we} offers a challenging VPR benchmark through data collected from a train-mounted camera along a fixed route across multiple seasonal changes; however, it lacks lidar data and provides no viewpoint diversity (\ie{} reverse revisits) due to the fixed trajectory of the train. 
Wild-Places~\cite{knights2022wild} and Oxford Forest~\cite{oh2024evaluation} provide large-scale benchmarks for LPR in forest environments with both intra-sequence and multi-session revisits, but they lack support for VPR or CMPR due to the absence of camera data.
RELLIS-3D~\cite{jiang2021rellis} includes camera and lidar recordings of several traversals of unpaved trails around a university campus, but the limited scale and diversity of its sequences restrict its utility as a place recognition benchmark.
While BotanicGarden~\cite{liu2024botanicgarden} provides synchronized image and lidar data from multiple traversals, its restriction to a single, small-scale botanical garden results in limited scene, viewpoint, and temporal diversity.
In contrast, \dsname{} introduces sequential RGB frames synchronized to the dense lidar submaps from the Wild-Places~\cite{knights2022wild} dataset for eight traversals across multiple natural environments, with long-term revisits and diverse viewpoints, establishing a challenging new benchmark for VPR and CMPR.\looseness=-1

\subsection{Metric Depth Estimation}

Metric depth estimation has recently made significant progress, driven by the emergence of 3D foundation models~\cite{yang2024depth,wang2025vggt}, which achieve strong generalization across diverse visual domains. While these advances have been transformative for computer vision, their impact is particularly critical in robotics, where reliable depth estimation underpins core capabilities such as SLAM~\cite{zhang2021survey,keetha2024splatam}, cross-modal place recognition~\cite{xu2024c2l}, and 3D structure-from-motion~\cite{wang2024dust3r,leroy2024grounding,wang2025vggt}.  Despite this progress, a key challenge in training metric depth estimation models is acquiring accurate ground-truth annotations in real-world environments. Synthetic datasets such as VirtualKITTI~\cite{gaidon2016virtual} can generate dense pixel-wise annotations across diverse conditions, yet models trained on them often struggle with the sim-to-real domain gap~\cite{yang2024depth}. Outdoor benchmarks such as KITTI Depth~\cite{Uhrig2017THREEDV} alleviate this issue by producing semi-dense depth maps through lidar accumulation, but these datasets only represent urban scenes.

In contrast, \dsname{} provides semi-dense sequential depth annotations in unstructured natural environments. By leveraging accumulated global point cloud maps and robust visibility estimation to remove occluded points (see Section~\ref{subsec:depthgeneration}), \dsname{} enables training and evaluation of metric depth estimation under the complex conditions of natural scenes, where irregular terrain and dense vegetation continue to challenge state-of-the-art models.

\section{WildCross Benchmark}
\vspace{1mm}
\dsname{} leverages the raw data from the Wild-Places~\cite{knights2022wild} LPR dataset and extends it into a cross-modal benchmark for place recognition and metric depth estimation through two main advances, complementary to WildScenes~\cite{vidanapathirana2025wildscenes}, which focuses on 2D and 3D semantic segmentation in the same natural environments.
Firstly, we reprocess the original traversals to produce sequential RGB frames at 15Hz with accurate 6DoF ground truth poses synchronized with dense 3D lidar submaps in the same environment.  Secondly, we introduce an annotation pipeline that generates semi-dense metric depth maps with surface normals for every RGB frame through leveraging the accumulated global point cloud and 6DoF camera poses for each image, using point visibility estimation to remove occluded points from each annotated frame. 
These contributions make \dsname{} a powerful benchmark for evaluating performance on the tasks of visual and cross-modal place recognition, in addition to metric depth estimation in complex unstructured natural environments.
Section~\ref{subsec:seqinfo} describes the sequences, while Sections~\ref{subsec:rgblidargen}, \ref{subsec:trajectorygen}, and \ref{subsec:depthgeneration} cover the generation of RGB and lidar data, the pose estimation, and the creation of semi-dense depth and surface normal annotations, respectively.

\subsection{Sequence Information}
\label{subsec:seqinfo}

For consistency with Wild-Places~\cite{knights2022wild} and WildScenes~\cite{vidanapathirana2025wildscenes}, we adopt the notation V-XX and K-XX to denote sequence XX on the Venman and Karawatha trajectories, respectively. Table~\ref{tab:sequence_information} summarizes the per-sequence and overall statistics of the dataset, including the number of submaps, image frames, intra-sequence revisits, and total traversal distance for each sequence. The traversals follow a consistent pattern: in both environments, Sequence 02 corresponds to the reverse trajectory of Sequence 01, Sequence 03 follows an alternate extended route, and Sequence 04 repeats the route of Sequence 01. As we show in Section~\ref{subsec:vpr}, state-of-the-art VPR approaches continue to face significant challenges in successfully relocalizing under reverse revisits.  The repeated traversals in \dsname{}, including reverse and alternate trajectories with substantial overlap, therefore provide a basis for evaluating both intra- and inter-sequence re-localization under challenging revisit conditions.\looseness=-1

\subsection{Color Images and Submap Generation}
\label{subsec:rgblidargen}
To obtain our color images, we extract camera frames from the raw video output of the forward-facing camera of the sensor payload at 15Hz.
These are then rectified using the distortion parameters obtained after sensor calibration.  
To generate our lidar submaps, we follow the approach of \cite{knights2022wild} and generate a global accumulated point cloud map with corresponding sensor trajectory calculated using a lidar-inertial SLAM~\cite{ramezani2022wildcat}.
We then use the SLAM trajectory to generate 3D submaps from the dense global map, sampling all points within a 30m radius and a 1s time window around the position of the sensor payload every 0.5s along the global trajectory.\looseness=-1

\subsection{Ground Truth Poses}
\label{subsec:trajectorygen}

For each image frame $\mathcal{I}$ in \dsname{} we provide ground-truth 6DoF poses in the world frame, represented as $T(t) = \{q(t), x(t)\}$, where $q(t) \in \mathbb{R}^4$ denotes rotation as a unit quaternion and $x(t) \in \mathbb{R}^3$ denotes the translation of the sensor origin at time $t$. To generate the poses, we interpolate between poses in the SLAM trajectory by applying spherical linear interpolation for rotation and linear interpolation for translation:\vspace{-2mm} 

\begin{equation}
T(t) = 
\left\{ \,
q(t) = q_2 \left(q_2^{-1}\cdot q_1 \right)^u , \; 
x(t) = (1-u)x_1 + u x_2
\right\}, 
\end{equation}

\begin{equation}
u = \frac{t - t_1}{t_2 - t_1},
\end{equation}
where $(q_1, x_1, t_1)$ and $(q_2, x_2, t_2)$ are the SLAM poses immediately before and after timestamp $t$.  Poses for the camera images are then transformed into camera coordinate frames using the extrinsic transform between the SLAM and camera frames, and the poses for multiple sessions in the same environment are aligned by using ICP to align the global point cloud maps produced by the SLAM for each sequence.\looseness=-1

\subsection{Metric Depth and Surface Normal Generation}
\label{subsec:depthgeneration}

\begin{table}[t!]
    \centering
    \addtolength{\tabcolsep}{-0.45em}
    \begin{NiceTabular}{ll |l| ccc| ccc}
        \hline 
         \Block{2-2}{Sequence} &&\Block{2-1}{Distance}& \Block{1-3}{Camera / Depth Images} &&& \Block{1-3}{Submaps} \\
         &&&  All & Revisits &\% & All & Revisits &\%\\
         \hline 
         \Block{4-1}{Venman} 
         & 01 &2.64km &35.3K    &10.6K & 30.03   &4.7K     &1.2K &  25.69\\
         & 02 &2.64km &34.1K    &8.3K  & 24.25   &4.6K   &980  &  21.51\\
         & 03 &4.59km &63.8K    &8.7K  & 13.70   &8.5K   &906  &  10.70\\
         & 04 &2.81km &43.1K    &8.0K  & 18.57   &5.7K   &937  &  16.33\\
          \hline 
         \Block{4-1}{Karawatha} 
         & 01 &5.14km &66.1K    &5.6K  & 8.43    &8.8K   &553  &  6.27\\
         & 02 &5.66km &75.6K    &6.6K  & 8.70    &10.0K   &522  &  5.18\\
         & 03 &6.27km &114.2K   &38.4K & 33.64   &15.2K  &2.5K &  16.73\\
         & 04 &3.17km &43.8K    &9.3K  & 21.35   &5.8K   &1.0K &  18.59\\
         \hline 
         \Block{1-2}{Total} 
         &&   33km&476K   & 95.5K & - & 63.3K       & 8.7K & - \\
    \end{NiceTabular} %
    \caption{Sequence statistics for \dsname{}. ``Revisits'' denote intra-sequence loop closures. \looseness=-1}
    \label{tab:sequence_information}
    \vspace{-2mm}
\end{table}

\begin{figure}[t!]
    \centering
    \includegraphics[width=\linewidth]{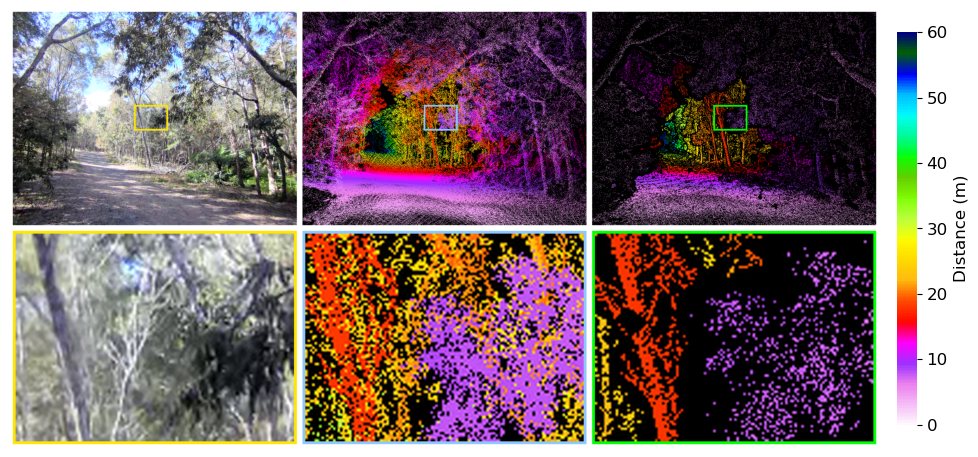}
    \begin{tabularx}{\columnwidth}{p{0.25\columnwidth}p{0.30\columnwidth}p{0.30\columnwidth}X}
        (a) RGB Image & (b) Noisy Depth & (c) Ours\\
    \end{tabularx}
    \caption{Impact of visibility estimation. (a) RGB Image, (b) Naïve projection of global 3D points produces noisy depth maps with occluded points.  (c) Our visibility pipeline removes these, yielding higher-quality depth.}
    \label{fig:badprojection}
    \vspace{-5mm}
\end{figure}

\begin{figure}[t]
    \centering
    \includegraphics[width=\columnwidth]{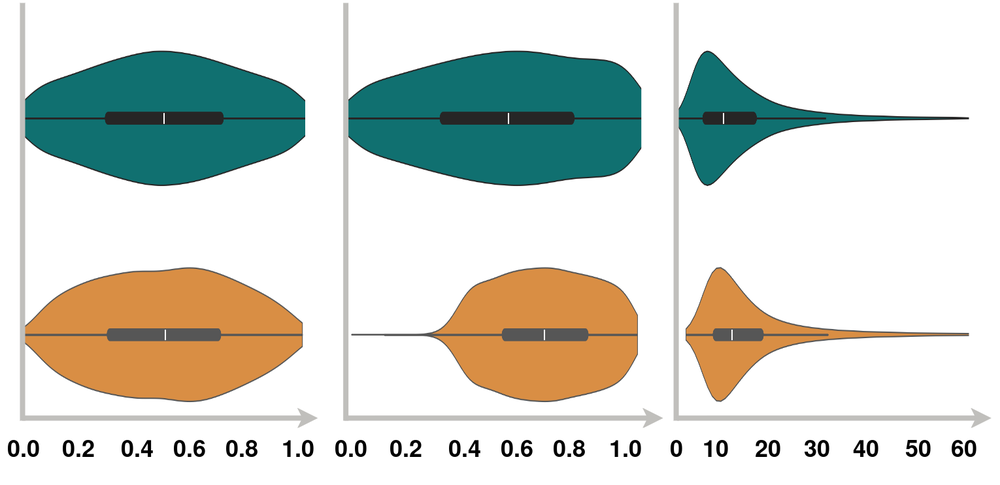}
\begin{tabularx}{\columnwidth}{p{0.008\textwidth}p{0.13\textwidth}p{0.12\textwidth}X}
        & (a) Width & (b) Height & (c) Depth (m)
    \end{tabularx}
    \caption{Depth distribution for \dsname{} (\textcolor{teal}{\large$\bullet$}) vs. KITTI Annotated Depth (\textcolor{orange}{\large$\bullet$})~\cite{Uhrig2017THREEDV}. Violin plots are computed from 1\% subsamples of both datasets. Width and height are normalized with respect to image sizes.\looseness=-1}
    \label{fig:violin}
    \vspace{-6mm}
\end{figure}

We provide sequential, semi-dense metric depth and surface normal annotations for each camera frame. Each pixel in the depth annotation, denoted as $D(u,v)$ at image coordinate $(u,v)$, represents the distance to the nearest object intersected by the ray originating from the camera center and passing through pixel $(u,v)$ in 3D space.
Likewise, the surface normal $SN(u,v)$ at image coordinate $(u,v)$ represents the surface normal vector at the same point with respect to the camera coordinate frame. 
Depth images are generated by projecting 3D points from the global point cloud onto the image plane using the camera projection function and the extrinsic transform relating the camera and lidar coordinate frames. Importantly, not all 3D points in the point cloud are visible in the captured image (see Figure~\ref{fig:badprojection}). Determining visibility is challenging because each 3D point has infinitesimal volume, making it unlikely for multiple points to lie along the same ray passing through the camera center.
To identify the visible points in an image,  we compute the surface normal of each point in the point cloud and select the points with a normal oriented toward the camera. Points that project outside the image bounds or lie behind the camera are also discarded. Finally, we eliminate points belonging to surfaces occluded by other surfaces.  The surface normal of each point $p_i \in  \mathbb{R}^3$ is estimated by computing the eigendecomposition of its local neighborhood, defined as all points within $0.5$m of $p_i$. 
The eigenvector corresponding to the smallest eigenvalue is selected as the surface normal, after ensuring it is oriented towards the observation location of the point $p_i$. These surface normals, oriented in the camera frame, are also used to generate the surface normal image after filtering of points for occlusions, as is done for depth. \looseness=-1 

To address the problem of occlusion, we utilize the generalized hidden point removal (GHPR) operator proposed in \cite{katz2015ovpc,vechersky2018paintcloud}. The operator consists of two steps: i) apply a spherical reflection to the filtered point cloud, which has the camera position as its origin, and ii) calculate the convex hull of the spherical reflected point cloud.  The spherical reflection of a 3D point $p \in \mathbb{R}^3$  is defined by the function $F(p; \gamma)$:\looseness=-1

\begin{align} F(p; \gamma) = \left \{ \begin{matrix} p \| p \|_2^{-1} f(\|p\|_2; \gamma) & \left\|p\right\|_2 \neq 0 \\ 0 &\left\|p\right\|_2 = 0,  \end{matrix} \right.  \label{eq:spherical-reflection-function} \end{align}
where the kernel function $f(d; \gamma)$ is a monotonically decreasing function of the distance $d > 0$. In this work, we use the exponential inversion kernel for its scale-invariant properties:\looseness=-1
\begin{align} f(d; \gamma) &= d^\gamma  & \gamma &< 0. \end{align}

The spherical reflection function $F$ has the property that a point which is close to the camera is transformed to a location that is far away from the camera, and vice versa. Therefore, we can determine which points are observed in the camera image by selecting those that lie on the convex hull of the spherically reflected point cloud.  We perform a statistical analysis comparing the depth data provided by \dsname{} to that of the annotated depth maps in the KITTI dataset~\cite{Uhrig2017THREEDV}, which were produced through the projection of accumulated lidar point clouds but in a structured urban environment.  We present the distribution of non-zero pixels from depth images as violin plots shown in Fig~\ref{fig:violin}.  As the datasets have different image sizes, the distributions of pixels across the image coordinates are provided in terms of relative coordinates within the image, while depth values are given in meters.  The most significant difference in the distribution between the datasets can be seen in the distribution of points along the height axis of the images, where the top portion of the image is unlabeled in the KITTI dataset due to the limited vertical Field-of-View (FoV) of its lidar, whereas our depth images provides a much denser depth along the height axis of the images which is crucial in natural environment. \looseness=-1

\begin{table}[t!]
    \centering
    \addtolength{\tabcolsep}{-0.4em}
    \begin{NiceTabular}{c|cccc|cc|cc}
         \Block{2-1}{Split} &\Block{2-1}{ V/K-01} & \Block{2-1}{V/K-02} & \Block{2-1}{V/K-03} & \Block{2-1}{V/K-04} & \Block{1-2}{Lidar} && \Block{1-2}{Camera} &   \\
         & &&&& Train&Test&Train&Test \\\hline 
         01 & \textbf{Test} & Train & Train & Train &49.8K & 13.5K& 374.6K & 101.4K \\
         02 & Train & \textbf{Test} & Train & Train &48.7K & 14.6K& 366.3K & 109.7K \\
         03 & Train & Train & \textbf{Test} & Train &39.7K & 23.6K& 298.0K & 177.9K \\
         04 & Train & Train & Train & \textbf{Test} &51.8K & 11.5K& 389.1K & 86.9K \\
         
    \end{NiceTabular}
    \caption{Train/test cross-fold splits for \dsname{}.}
    \label{tab:splits}
    \vspace{-6mm}
\end{table}

\section{Experiments}
\label{sec:experiments}
\subsection{Training and Testing Splits}
We observed that the training splits introduced for LPR in Wild-Places~\cite{knights2022wild} are not well suited to training large-scale VPR and CMPR networks. To address this, we propose a cross-fold training and evaluation setup for benchmarking VPR and CMPR performance on \dsname{}. In this setup, training and evaluation follow a four-fold cross-split design. In each split, the sequences with the same index (\eg{} Split-1, V-01, and K-01) from both environments are held out together for evaluation, while the remaining sequences are used for training. 
Table \ref{tab:splits} reports the number of training and testing samples in each split. During evaluation, the held-out sequences are used for intra-sequence place recognition and also serve as queries for inter-sequence recognition, with the training sequences acting as the database.\looseness=-1

\subsection{Visual Place Recognition (VPR)}
\label{subsec:vpr_exp}

\begin{table*}[t]
    \centering
    \addtolength{\tabcolsep}{-0.25em}
    \resizebox{\textwidth}{!}{
    \begin{NiceTabular}{ll | cccccccccccccccc|cc}
         \Block{2-2}{Method} && 
         \Block{1-2}{V-01} && \Block{1-2}{V-02} && \Block{1-2}{V-03} && \Block{1-2}{V-04} &&
         \Block{1-2}{K-01} && \Block{1-2}{K-02} && \Block{1-2}{K-03} &&\Block{1-2}{K-04} &&\Block{1-2}{Average}\\
         && R1 & R5 & R1 & R5 & R1 & R5 &R1 & R5 
         & R1 & R5 & R1 & R5 & R1 & R5 & R1 & R5 & R1 & R5 \\
        \hline 
        \Block{4-1}<\rotate>{Zero-shot}
        &NetVLAD~\cite{arandjelovic2016netvlad} &47.24&51.91&\textbf{52.44}&\textbf{64.70}&7.040&14.65&38.98&47.91&47.09&59.10&55.70&73.86&13.76&23.34&51.98&55.67&39.28&48.89\\
        &MixVPR~\cite{ali2023mixvpr}            &51.86&57.39&44.60&50.63&7.33&13.80&44.60&50.84&74.89&81.88&\textbf{69.53}&\textbf{85.18}&24.88&32.34&\textbf{57.12}&\textbf{59.58}&\textbf{46.85}&\textbf{53.96}\\
        &SALAD~\cite{Izquierdo_CVPR_2024_SALAD} &50.64&57.01&44.60&49.06&\textbf{13.05}&\textbf{19.58}&43.72&48.53&78.48&83.50&54.71&68.16&\textbf{26.67}&\textbf{32.78}&54.93&59.31&45.85&52.24\\
        &BoQ~\cite{ali2024boq}                  &\textbf{54.65}&\textbf{62.48}&48.22&55.88&11.62&17.69&\textbf{46.66}&\textbf{52.47}&\textbf{80.99}&\textbf{84.22}&45.06&55.62&26.61&31.36&55.41&58.19&46.15&52.24\\
        \hline 
        \Block{4-1}<\rotate>{Fine-tuned}
        &NetVLAD~\cite{arandjelovic2016netvlad} &68.95&72.16&69.89&74.05&18.09&22.72&59.90&66.02&83.86&87.53&86.32&91.57&26.82&32.35&56.32&59.96&58.77&63.30\\
        &MixVPR~\cite{ali2023mixvpr}            &66.16&68.52&68.86&72.78&17.80&25.01&52.90&57.34&84.66&87.89&87.99&89.59&30.40&38.22&60.87&\textbf{65.04}&58.71&63.05\\
        &SALAD~\cite{Izquierdo_CVPR_2024_SALAD} &69.04&72.68&72.24&\textbf{80.33}&23.30&29.48&58.96&64.65&86.37&89.33&88.98&91.26&\textbf{34.05}&\textbf{38.83}&\textbf{61.13}&62.53&61.76&66.14\\
        &BoQ~\cite{ali2024boq}                  &\textbf{70.22}&\textbf{74.61}&\textbf{72.84}&77.67&\textbf{28.68}&\textbf{38.87}&\textbf{62.40}&\textbf{68.71}&\textbf{87.62}&\textbf{89.51}&\textbf{90.81}&\textbf{92.93}&32.26&38.05&60.49&62.90&\textbf{63.17}&\textbf{67.91}\\
    \end{NiceTabular}
    }
    \caption{Intra-sequence VPR results on \dsname{} for zero-shot and fine-tuned networks.}
    \label{tab:vpr_intra}
    \vspace{-4mm}
\end{table*}

\begin{table}[t]
    \centering
    \addtolength{\tabcolsep}{-0.35em}
    \begin{NiceTabular}{ll|cccc|cc}
         \Block{2-2}{Method (Backbone)} && \Block{1-2}{Venman} && \Block{1-2}{Karawatha} && \Block{1-2}{Average}  \\
         && R1 & R5 & R1 & R5 & R1 & R5 \\
         \hline 
         \Block{4-1}<\rotate>{Zero-Shot} 
         &NetVLAD~\cite{arandjelovic2016netvlad}    &25.86&43.94&16.15&29.73&21.00&36.84\\
         &MixVPR~\cite{ali2023mixvpr}               &54.10&61.41&35.73&44.31&44.92&52.86 \\
         &SALAD~\cite{Izquierdo_CVPR_2024_SALAD}    &57.49&64.49&41.27&50.14&49.38&57.32\\
         &BoQ~\cite{ali2024boq}                     &\textbf{61.62}&\textbf{67.98}&\textbf{45.89}&\textbf{54.98}&\textbf{53.76}&\textbf{61.48} \\
         \hline
         \Block{4-1}<\rotate>{Fine-tuned}
         &NetVLAD~\cite{arandjelovic2016netvlad}    &64.31&67.49&46.94&52.43&55.63&59.96\\
         &MixVPR~\cite{ali2023mixvpr}               &65.30&68.58&50.24&55.80&57.77&62.19\\
         &SALAD~\cite{Izquierdo_CVPR_2024_SALAD}    &68.54&71.86&54.29&59.86&61.41&65.86\\
         &BoQ~\cite{ali2024boq}                     &\textbf{68.66}&\textbf{72.01}&\textbf{55.07}&\textbf{60.37}&\textbf{61.87}&\textbf{66.19}\\
    \end{NiceTabular}
    \caption{Inter-Sequence VPR Results on \dsname{} for zero-shot and fine-tuned networks. }
    \label{tab:vpr_inter}
    \vspace{-5mm}
\end{table}

We evaluate four state-of-the-art methods for VPR: 
NetVLAD \cite{arandjelovic2016netvlad}, MixVPR~\cite{ali2023mixvpr}, DINOv2-SALAD (SALAD)~\cite{Izquierdo_CVPR_2024_SALAD}, and Bag-of-Queries (BoQ)~\cite{ali2024boq}.  
Unlike LPR, positive training pairs for VPR cannot be formed solely using a distance threshold. The limited field-of-view of the camera and the presence of reverse revisits within and across sequences can result in false positives, where two images selected as a pair share little or no visual overlap. To mitigate this, and inspired by ~\cite{Berton_CVPR_2022_CosPlace}, we define positive training pairs in \dsname{} as images whose camera poses are within 5m distance and 15\degree~bearing of each other, and negative pairs are those separated by more than 50m. At evaluation, a retrieved image is considered a correct match if its pose lies within 25m of the query. 
We report results under two evaluation settings: zero-shot and fine-tuned. In the \textbf{zero-shot} setting, each method is evaluated on \dsname{} using its best released pretrained model, without any additional training on \dsname{}. This setup reflects the scenario where models pretrained on large-scale, predominantly urban, datasets are applied directly to natural environments, thereby measuring their out-of-domain generalization ability. 

In the \textbf{fine-tuned} setting, the same methods are further fine-tuned on the \dsname{} training splits before evaluation. This measures their in-domain performance once exposed to data from natural environments. Reporting both settings highlights the gap between cross-domain generalization (urban-to-natural) and in-domain adaptation, providing a comprehensive benchmark of VPR in natural environments.\looseness=-1

\subsection{Cross-Modal Place Recognition (CMR)}

Cross-modal place recognition (CMPR) aims to localize across different sensing modalities, such as retrieving lidar submaps given visual queries. This task is particularly challenging in natural environments, where structural complexity and viewpoint variation exacerbate the difficulty of aligning cross-modal features. To explore this task on \dsname{}, we evaluate LIP-Loc~\cite{shubodh2024lip}, a baseline learning-based CMPR method. Within our cross-fold training and testing regime, we evaluate the inter-sequence CMPR performance using the images from the unseen sequences as queries and lidar submaps from all sequences in an environment as databases.  
When trained with the original batched contrastive loss from~\cite{shubodh2024lip} we observed a collapse in the feature space, which we attribute to the lack of any mechanisms in the LIP-Loc loss to deal with false negative examples in the batch (\ie{} negative samples with high real-world similarity). To address this limitation, we replace the contrastive formulation with a modified cross-entropy loss as follows: 
\begin{equation}
    \mathcal{L}_{\text{LIP-Loc}} = \mathcal{L}_{CE}\left(
        \frac{
        \exp\left(v_i^\top v_i^+\right)}
        {\exp\left(v_i^\top v_i^+\right)+\sum_{j\notin nn_i}\exp\left(v_i^\top v_j\right)}
        \right),
\end{equation}
where $v_i$, $v_i^+$, $v_j$ represent global feature vectors for a training sample and its corresponding positive and negative samples in the batch respectively, $nn_i$ denotes the set of non-negatives in the batch for $v_i$ (\ie{} samples located within 50m in the real world), and $\mathcal{L}_{CE}\left(x\right)=-x\log\left(x\right)\left(x\right)=-x\log\left(x\right)$ is the cross-entropy loss.  We use a positive threshold of 25m during evaluation. 
While the original LIP-Loc framework employed ResNet50 on ImageNet-pretrained visual transformers as its backbone, we extend this setup by incorporating more powerful recent pretraining approaches. Specifically, we evaluate DINOv2~\cite{oquab2023dinov2} and DINOv3~\cite{simeoni2025dinov3}, using the ViT-S encoder and extracting the class token, followed by a fully connected layer, as the global feature representation.
Note, we report the best configuration results found experimentally for each method, with DINOv2 having a frozen backbone and DINOv3 having an unfrozen backbone when fine-tuning.

\subsection{Lidar Place Recognition (LPR)}
\label{subsec:lpr}

To facilitate comparison with purely 3D place recognition approaches, we evaluate three learning-based LPR methods: MinkLoc3Dv2~\cite{komorowski2022improving}, LoGG3D-Net~\cite{vidanapathirana2022logg3d}, and HOTFormerLoc~\cite{griffiths2025hotformerloc}.
For training, we define positive and negative pairs using distance thresholds of 3m and 20m, respectively. At evaluation, a retrieved submap is considered a correct match if its pose is within 3m of the query.

\subsection{Metric Depth Estimation}
\dsname{} also supports training and evaluation for metric depth estimation. We evaluate this task using DepthAnythingV2~\cite{yang2024depth} as a representative state-of-the-art baseline. For this experiment, sequences V-01 and K-01 are held out for testing, K-02 is used for validation, and the remaining sequences are used for training.
We report three common metrics:
threshold accuracy ($\delta_1$), which measures the percentage of the predicted pixels whose depth differs from ground truth by no more than 25\%, Absolute Relative Error (AbsRel), which quantifies the average relative difference between predicted and true depths, and Root Mean Square Error (RMSE), which measures the overall deviation of predictions from the ground truth. As with VPR, we report results under both zero-shot and fine-tuned settings. In the zero-shot setting, we directly evaluate the released model trained on KITTI~\cite{geiger2013vision} and VirtualKitti~\cite{gaidon2016virtual}, thereby assessing Out-Of-Domain (OOD) generalization from urban to natural environments. In the fine-tuned setting, we adapt the model to \dsname{} depth images, measuring its in-domain performance. %

\section{Results}
\label{sec:results}

\subsection{Visual Place Recognition}
\label{subsec:vpr}

\begin{figure}[t!]
    \centering
    \begin{subfigure}[b]{0.48\columnwidth}
         \centering
         \includegraphics[width=\textwidth]{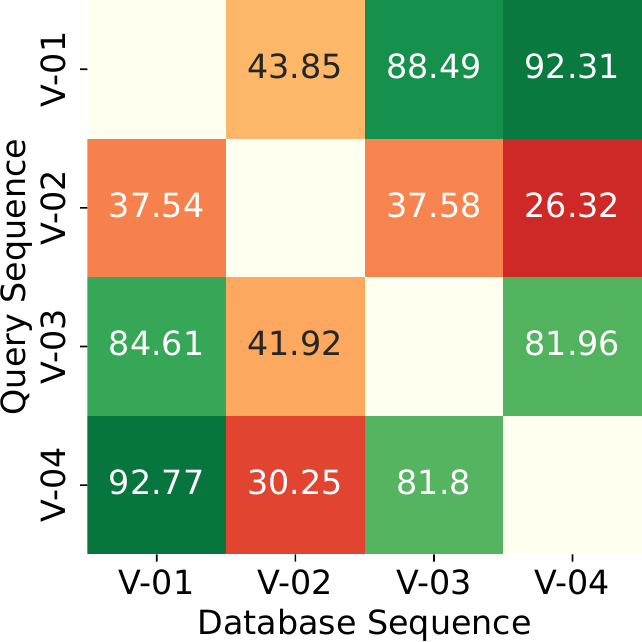}
         \caption{Venman, Zero-Shot}
     \end{subfigure}
     \hfill 
     \begin{subfigure}[b]{0.48\columnwidth}
         \centering
         \includegraphics[width=\textwidth]{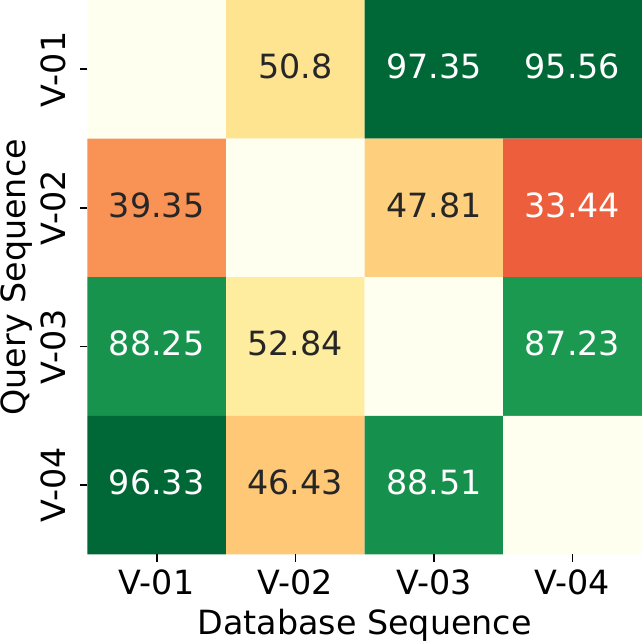}
         \caption{Venman, Fine-Tuned}
     \end{subfigure}
     
     \begin{subfigure}[b]{0.48\columnwidth}
         \centering
         \includegraphics[width=\textwidth]{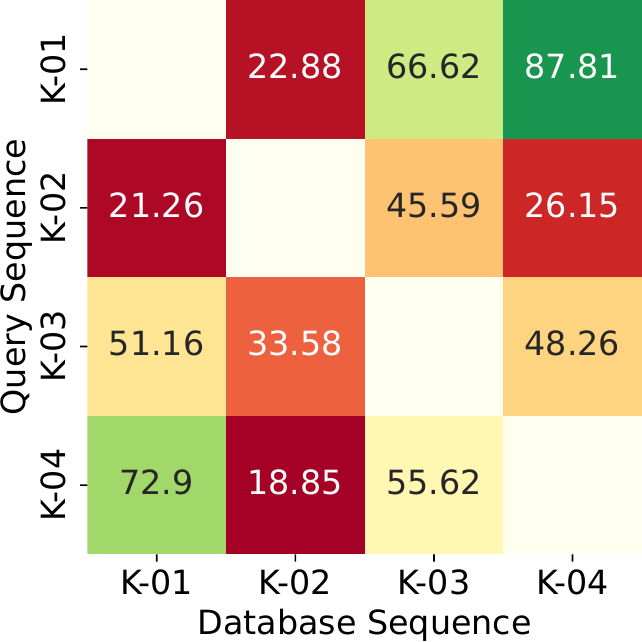}
         \caption{Karawatha, Zero-Shot}
     \end{subfigure}
     \hfill 
    \begin{subfigure}[b]{0.48\columnwidth}
         \centering
         \includegraphics[width=\textwidth]{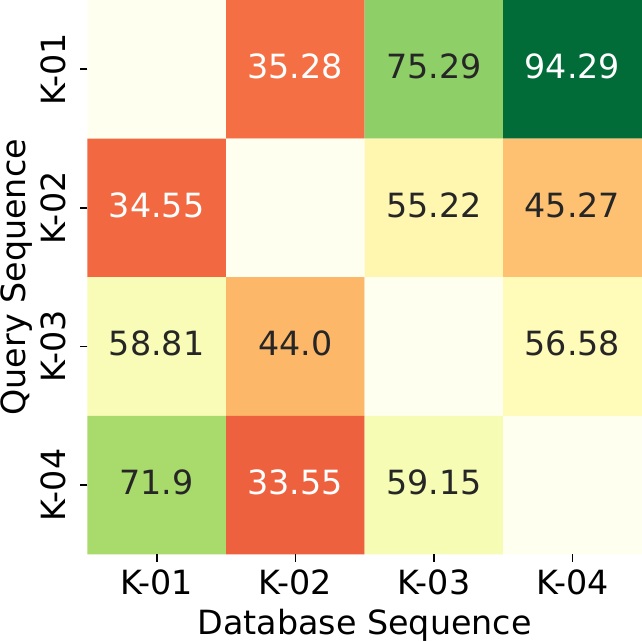}
         \caption{Karawatha, Fine-Tuned}
     \end{subfigure}
     \caption{Cross-sequence VPR R1 on \dsname{}. Reverse revisit sequences V-02 and K-02 significantly degrade performance, underscoring the challenge of viewpoint diversity. \looseness=-1}

     \label{fig:vpr_inter_breakdown}
     \vspace{-6mm}
\end{figure}

Tables~\ref{tab:vpr_intra} and~\ref{tab:vpr_inter} summarize intra- and inter-sequence VPR performance on \dsname{}. Fine-tuning consistently improves performance across all methods, showcasing the value of training on in-domain natural environment data. Nevertheless, even the strongest overall method, BoQ~\cite{ali2024boq}, achieves only 64.45\% R1 for intra-sequence and 61.44\% for inter-sequence evaluation. In comparison, on established urban benchmarks such as Pittsburgh~\cite{torii2013visual} and MSLS~\cite{warburg2020mapillary}, the same method exceeds 90\% R1.\looseness=-1

\begin{table*}[t]
    \centering
    \addtolength{\tabcolsep}{-0.35em}
    \resizebox{\textwidth}{!}{
    \begin{NiceTabular}{l | cccccccccccccccc|cc}
         \Block{2-1}{Method} & 
         \Block{1-2}{V-01} && \Block{1-2}{V-02} && \Block{1-2}{V-03} && \Block{1-2}{V-04} &&
         \Block{1-2}{K-01} && \Block{1-2}{K-02} && \Block{1-2}{K-03} &&\Block{1-2}{K-04} &&\Block{1-2}{Average}\\
         & R1 & R5 & R1 & R5 & R1 & R5 &R1 & R5 
         & R1 & R5 & R1 & R5 & R1 & R5 & R1 & R5 & R1 & R5 \\
        \hline 
        MinkLoc3Dv2~\cite{komorowski2022improving} & 96.53&\textbf{100.0}&96.02&\textbf{100.0}&59.27&93.27&89.11&99.04&71.97&98.73&\textbf{98.28}&\textbf{100.00}&\textbf{57.80}&82.68&86.84&99.35&81.98&96.64\\
        {LoGG3D-Net}~\cite{vidanapathirana2022logg3d} & 94.54&99.75&94.08&99.69&\textbf{83.22}&\textbf{97.57}&86.34&95.09&\textbf{98.55}&\textbf{99.82}&95.02&98.85&57.28&\textbf{93.66}&63.30&92.40&84.04&\textbf{97.10} \\
        HOTFormerLoc~\cite{griffiths2025hotformerloc} &\textbf{97.35}&99.67&\textbf{96.84}&99.90&50.88&84.55&\textbf{92.85}&\textbf{99.47}&97.29&99.64&97.32&\textbf{100.00}&56.50&68.58&\textbf{95.46}&\textbf{99.54}&\textbf{85.56}&93.92\\
    \end{NiceTabular}
    }
    \caption{Intra-sequence LPR results on \dsname{}. }
    \label{tab:lpr_intra}
    \vspace{-3mm}
\end{table*}

\begin{table}[t]
    \centering
    \addtolength{\tabcolsep}{-0.35em}
    \begin{NiceTabular}{l|cccccc}
         \Block{2-1}{Method} & \Block{1-2}{Venman} && \Block{1-2}{Karawatha} && \Block{1-2}{Average}  \\
         & R1 & R5 & R1 & R5 & R1 & R5 \\
         \hline 
         MinLoc3Dv2~\cite{komorowski2022improving}      &90.67&98.88&\textbf{78.58}&\textbf{92.33}&84.62&\textbf{95.61} \\
         LoGG3D-Net~\cite{vidanapathirana2022logg3d}    &84.39&94.19&72.09&86.02&78.24&90.11 \\
         HOTFormerLoc~\cite{griffiths2025hotformerloc}  &\textbf{91.60}&\textbf{99.09}&78.11&91.62&\textbf{84.85}&95.36 \\
    \end{NiceTabular}
    \caption{Inter-Sequence LPR Results on \dsname{}.}
    \label{tab:lpr_inter}
\end{table}

\begin{table}[t]
    \centering
    \addtolength{\tabcolsep}{-0.35em}
     \begin{NiceTabular}{l|cccccc}
         \Block{2-1}{Method (Backbone)} & \Block{1-2}{Venman} && \Block{1-2}{Karawatha} && \Block{1-2}{Average}  \\
         & R1 & R5 & R1 & R5 & R1 & R5 \\
         \hline 
         LIP-Loc (ResNet50) &40.16&54.45&34.25&48.91&37.20&51.68 \\
         LIP-Loc* (DINOv2-s) &52.55&62.71&45.26&\textbf{57.40}&48.90&60.06\\
         LIP-Loc* (DINOv3-s) &\textbf{56.54}&\textbf{63.19}&\textbf{48.16}&57.06&\textbf{52.35}&\textbf{60.12}\\
    \end{NiceTabular}
    \caption{Cross-Modal Place Recognition Results on \dsname{}.  * ViT-S for the pretrained model backbone.}
    \label{tab:cmpr_results}
    \vspace{-6mm}
\end{table}

\begin{table}[t]
    \centering
    \begin{NiceTabular}{ll|ccc}
         & Method (Backbone) & $\delta_1\uparrow$ & AbsRel$\downarrow$ & RMSE$\downarrow$\\
         \hline 
         \Block{3-1}{Zero-Shot} 
         & DepthAnythingV2 (ViT-S) &\textbf{0.284}&\textbf{0.558}&\textbf{7.651} \\
         & DepthAnythingV2 (ViT-B) &0.222&0.769&7.915 \\
         & DepthAnythingV2 (ViT-L) &0.074&1.478&13.734\\
         \hline 
         \Block{3-1}{Fine-tuned} 
         &  DepthAnythingV2 (ViT-S) &0.746&0.172&3.412 \\
         &  DepthAnythingV2 (ViT-B) &0.766&0.167&3.289 \\
         &  DepthAnythingV2 (ViT-L) &\textbf{0.789}&\textbf{0.157}&\textbf{3.150} \\
    \end{NiceTabular}
    \caption{Zero-Shot and Fine-tuned DepthAnythingV2~\cite{yang2024depth} results on \dsname{}.}
    \label{tab:depthpred}
    \vspace{-2mm}
\end{table}

One of the primary challenges in our benchmark derives from the prevalence of reverse revisits in both intra- and inter-sequence evaluation. As shown in Figure~\ref{fig:vpr_inter_breakdown}, inter-sequence Recall@1 drops substantially when K-02 or V-02 appear as either query or database sequences; these correspond to the reverse trajectories described in Section~\ref{subsec:seqinfo}. Similarly, Table~\ref{tab:vpr_intra} shows that the sequences with the highest number of intra-sequence reverse revisits (V-03 and K-03) also yield the lowest performance, with the best method achieving only 28.68\% and 36.57\% R1, respectively.
These results highlight that unstructured natural environments, particularly reverse revisits, remain a persistent challenge for state-of-the-art VPR methods. This highlights the value of \dsname{} as a benchmark for advancing visual place recognition under these difficult and underexplored conditions.\looseness=-1

\subsection{Lidar Place Recognition}

Tables~\ref{tab:lpr_intra} and~\ref{tab:lpr_inter} present intra- and inter-sequence LPR results on the new cross-fold splits introduced in \dsname{}. For intra-sequence evaluation, state-of-the-art methods achieve strong performance, with R1 scores above 90\% on average. This indicates that LPR methods can effectively handle revisits within the same sequence in natural environments.  In contrast, inter-sequence evaluation remains more challenging, with all LPR methods achieving average R1 scores below 86\%. This shows that long-term and multi-session generalization in unstructured environments remains challenging.\looseness=-1

\subsection{Cross-Modal Place Recognition}

Table~\ref{tab:cmpr_results} reports CMPR results on \dsname{}. Performance across all configurations remains limited, with the best result obtained by LIP-Loc using DINOv3 pretraining, which achieves an average R1 score of 51.42\%. This indicates that cross-modal retrieval in unstructured natural environments is particularly challenging. 

We also observe that the choice of backbone has a substantial effect on performance. Substituting the ImageNet-pretrained ResNet50 with transformer-based backbones yields consistent improvements, with DINOv3 providing an approximately 15\% increase in average R1 compared to the ResNet50 baseline. While this trend reflects the benefits of stronger visual pretraining, the overall performance remains considerably below that observed in VPR and LPR tasks. Progress will likely require approaches that explicitly address the domain gap between 2D image features and 3D structural representations, rather than relying on backbone improvements alone. By providing large-scale cross-modal data together with depth ground truth, \dsname{} offers the basis for training and evaluating more complex models that can better exploit cross-modal information for place recognition in natural environments.

\subsection{Metric Depth Estimation}

\begin{figure}
    \centering
    \begin{subfigure}[b]{0.48\columnwidth}
         \centering
         \includegraphics[height=3cm]{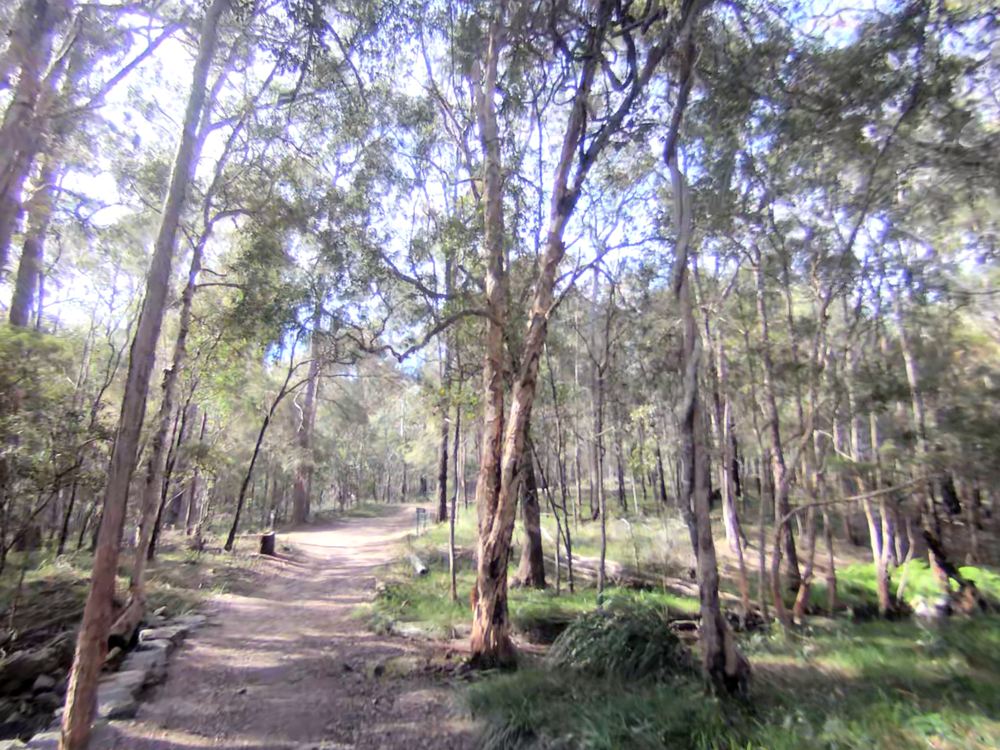}
         \caption{RGB Image}
     \end{subfigure} \hspace{-2mm}
     \begin{subfigure}[b]{0.48\columnwidth}
        
         \centering
         \includegraphics[height=3cm]{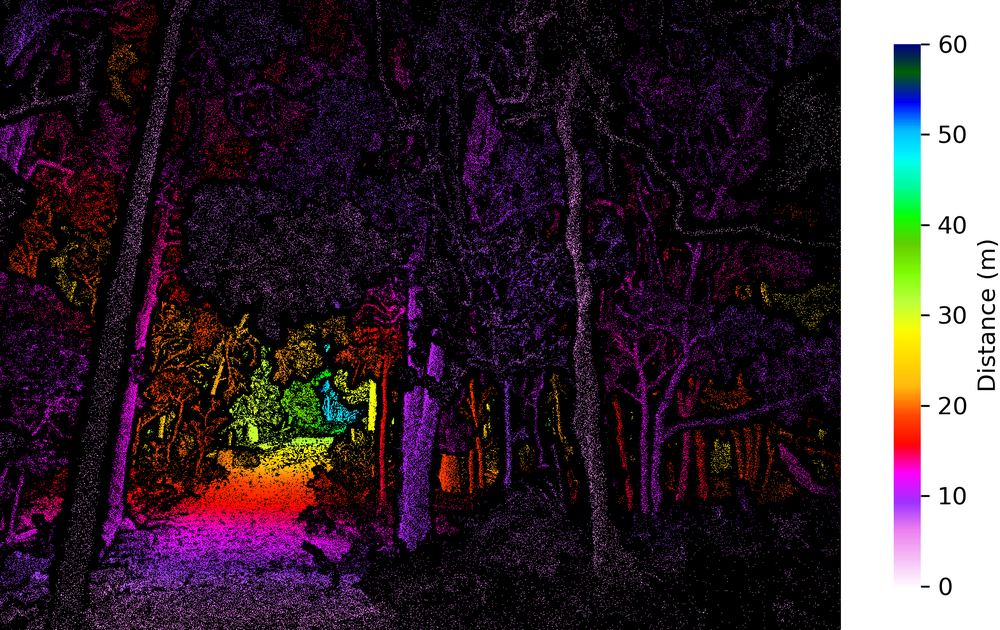}
         \caption{Ground Truth Depth}
     \end{subfigure}
     
     \begin{subfigure}[b]{0.48\columnwidth}
         \centering
         \includegraphics[height=3cm]{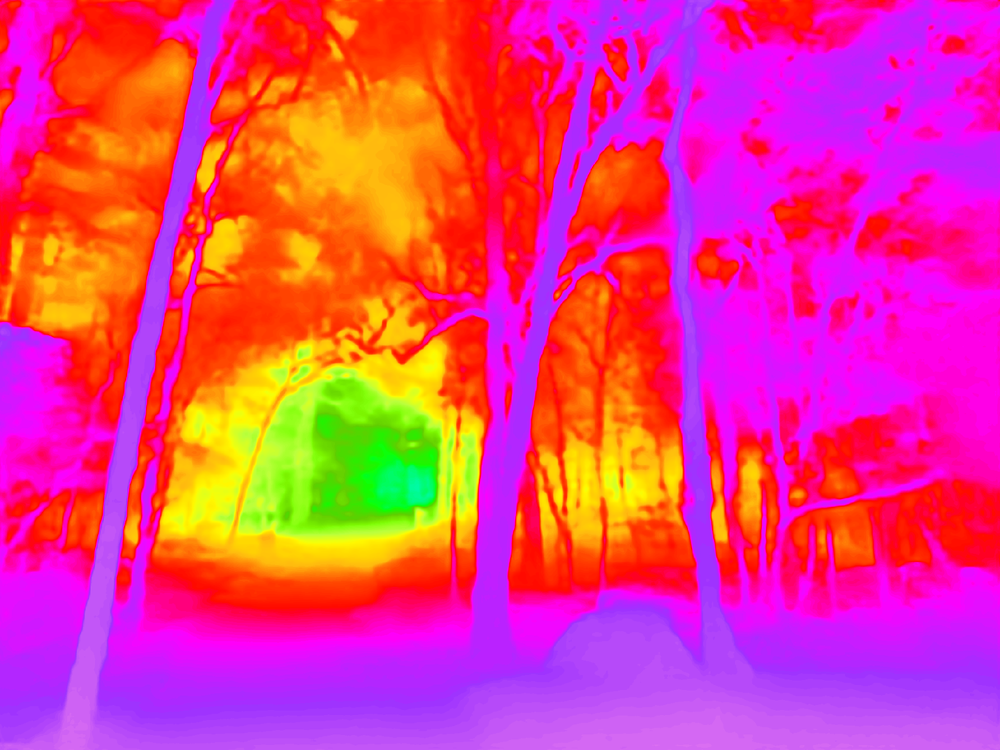}
         \caption{Zero-Shot Depth Prediction}
     \end{subfigure} \hspace{-2mm}
     \begin{subfigure}[b]{0.48\columnwidth}
        
         \centering
         \includegraphics[height=3cm]{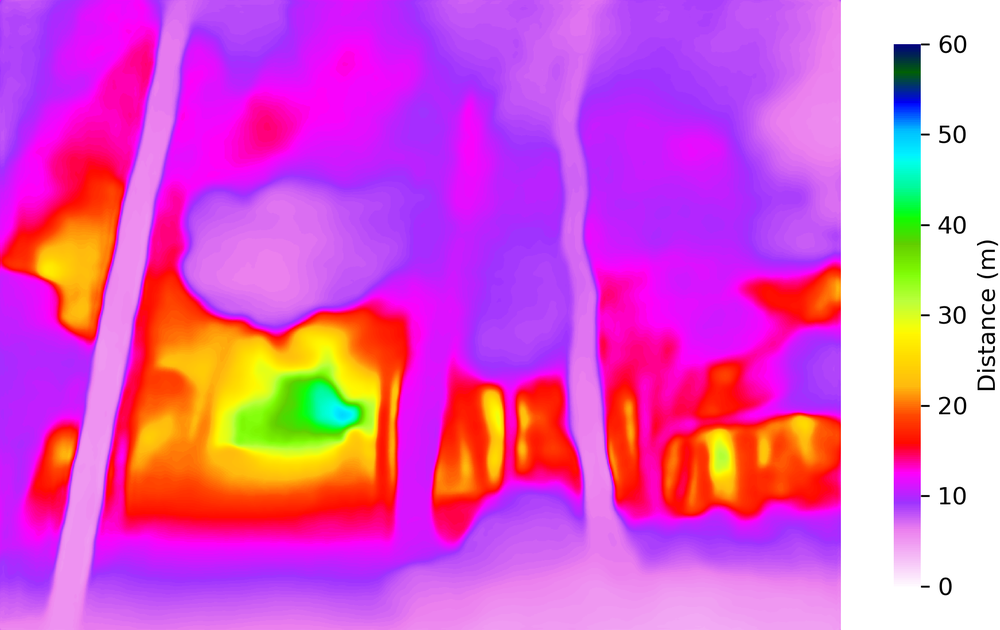}
         \caption{Fine-tuned Depth Prediction}
     \end{subfigure}

    \caption{Zero-shot vs. fine-tuned depth predictions. Fine-tuning improves scale but reduces detail.}%
    \label{fig:depthvis}
    \vspace{-5mm}
\end{figure}

Table~\ref{tab:depthpred} reports zero-shot and fine-tuned metric depth prediction results on \dsname{}. Fine-tuning with our depth annotations consistently improves the performance of DepthAnythingV2~\cite{yang2024depth} across all backbones, with larger ViT backbones yielding stronger results. 
In the zero-shot setting, however, larger models perform worse, with RMSE increasing by more than 5m from ViT-S to ViT-L.  %
Qualitative results in Figure~\ref{fig:depthvis} show that fine-tuning improves the overall scale of the predictions but reduces fine-grained detail compared to the pretrained model. Adapting depth prediction methods to the high-frequency structure of natural environments, characterized by foliage, irregular terrain, and fine textures, remains a key challenge compared to urban scenes dominated by walls and planar surfaces. 

Beyond spatial accuracy, temporal consistency is critical for robotics, as frame-to-frame flickering undermines reliable perception and navigation. To the best of our knowledge, \dsname{} is the first large-scale benchmark with sequential depth ground truth in natural environments, providing a foundation for systematic evaluation and advancement of temporally consistent metric depth estimation for robotic systems.\looseness=-1

\section{Conclusion}

In this paper, we introduced \dsname{}, a large-scale benchmark for cross-modal place recognition and metric depth estimation in natural environments. The dataset comprises over 476K sequential RGB frames with semi-dense depth and surface normal annotations, each aligned with accurate 6DoF poses and synchronized dense lidar submaps. Alongside describing an annotation pipeline for generating sequential depth ground truth, we evaluated a range of state-of-the-art methods for visual, lidar, and cross-modal place recognition, as well as metric depth estimation, and illustrated that even leading approaches struggle under these conditions.
By capturing the complexity of natural environments, dense vegetation, irregular terrain, and diverse viewpoints, \dsname{} highlights open research gaps where current methods struggle. In particular, reverse revisits in place recognition and temporally consistent depth estimation remain unresolved problems that are central to robust robotic autonomy. We hope \dsname{} inspires future research into bridging 2D and 3D perception and developing methods capable of reliable operation in complex natural environments.
\looseness=-1

\section*{Acknowledgment}
The authors would like to acknowledge support from the CRC-P Round 16 in partnership with Emesent. 

\balance{}

\bibliographystyle{./IEEEtran}
\bibliography{ref}

\end{document}